# Can empathy affect the attribution of mental states to robots?

CRISTINA GENA, Department of Computer Science, University of Torino, Italy

FRANCESCA MANINI, University of Torino, Italy

ANTONIO LIETO, Department of Computer Science, University of Torino, Italy

ALBERTO LILLO, Department of Computer Science, University of Torino, Italy

FABIANA VERNERO, Department of Computer Science, University of Torino, Italy

This paper presents an experimental study showing that the humanoid robot NAO, in a condition already validated with regards to its capacity to trigger situational empathy in humans, is able to stimulate the attribution of mental states towards itself. Indeed, results show that participants not only experienced empathy towards NAO, when the robot was afraid of losing its memory due to a malfunction, but they also attributed higher scores to the robot emotional intelligence in the Attribution of Mental State Questionnaire, in comparison with the users in the control condition. This result suggests a possible correlation between empathy toward the robot and humans' attribution of mental states to it.

## 1 INTRODUCTION

According to Preston and De Waal [33] empathy can be defined as "the capacity to (a) be affected by and share the emotional state of another, (b) assess the reasons for the others' state, and (c) identify with the other, adopting his or her perspective". Following a shared categorization in psychology [29, 31], empathy can be divided in three major categories: (1) empathy as an affective response to others' emotional states (*affective empathy*), (2) empathy as the cognitive understanding of others' emotional states, as well as the ability to put oneself in the other person's shoes (*cognitive empathy*), and (3) empathy as composed of both an affective and a cognitive component. Other perspectives [13, 39, 40] distinguish empathy in *dispositional empathy* and *situational empathy*. While the former is a character trait, namely a person's general tendency to empathize, the latter is the empathy that a human perceives towards another agent in a specific situation.

Indeed, empathy is a concept that affects multiple fields of knowledge, from social to developmental, from clinical psychology to neuroscience. Since the discovery in 1996 of mirror neurons [14], interest in the concept of empathy has increased exponentially, also involving the field of human-robot interaction, see for instance [21, 23, 31, 38]. Similarly, during a human-robot interaction, we speak of the cognitive process when a robotic agent appears to individuals as being able to understand and imitate the emotions of others. The affective process occurs when the robotic agent manifests its emotions through voice, body posture, movements and gestures, adapted to the context of the situation.

Several experiments have been conducted over time to study empathy in human-robot interaction, and will be described in the following Section.

According to several neurological and psychological researches [4, 17, 35] the involvement of mirror neuron system is implicated in neurocognitive functions, such as social cognition, language, *empathy*, and *Theory of Mind (ToM)* [3, 44], which is a human-specific ability that allows the attribution of mental states –intentions, thoughts, desires, and emotions– to themselves and others to explain and predict behavior. In particular, the attribution of mental states (AMS) has been defined as "the cognitive capacity to reflect upon one's own and other persons' mental states such as beliefs, desires, feelings and intentions" [2]. In everyday human-to-human interactions, such attributions are ubiquitous, although we are typically not necessarily aware of the fact that they are attributions —or the fact that they are attributions of mental states. In the attribution of mental states to others, human and nonhuman, empathy may have a key role, in particular considering constructs such as understanding the perspective of others, which is part of the previously introduced cognitive empathy.

This paper presents an experimental study showing that the humanoid and social robot NAO, in an already validated situation, is able to trigger empathy in humans and that such empathic response impacts participants' beliefs about the robot's capability of experiencing emotions and, in general, their perception of the robot's mental qualities, namely their attribution of mental states to the robot. This experiment was inspired by the study conducted by Seo et al. [37] in 2015, which investigated situational empathy, that is, the empathy that a human perceives towards another agent -in this case, a robot- in a specific situation and, in particular, when a sudden and unpleasant event arises. However, our experimental goals were slightly different: on the one hand, following Seo et al. [37], we wanted to ascertain whether an unexpected event, i.e., a functional problem in the NAO robot, and the consequent emotions of fear displayed by the robot, could trigger situational empathy ($H_2$), making participants feel sorry for NAO, in the eventuality of a loss of memory. On the other hand, we wanted to understand if the robot's ability to display -and, possibly, elicit- emotions, namely empathy, could impact participants' beliefs about the robot's ability to experience emotions ($H_1$), as well as their assumptions regarding the attribution of mental states to the robot ($H_3$).

Our results confirm that participants experienced empathic emotions towards NAO when the robot was feeling bad, thus implying that the so called situational empathy was successfully elicited, as expected based on the research of Seo et al. [37]. We also administered to the participants the *Attribution of Mental States Questionnaire* (AMS-Q) [27], which is considered suitable to be easily administered and sensitive for assessing the attribution of mental and sensory states to humans and nonhuman agents. Our results show that not only subjects in the experimental situation empathized with the robot, but also attributed higher emotional intelligence to the robot than the subjects in the control group, suggesting that its ability to display and elicit emotions made it appear as more "human". These results are very promising and suggest a connection between empathy and mental state attribution also in the context of Human Robot Interaction (HRI). To the authors' knowledge, there are no other experiments that have explicitly linked these two aspects in the field of HRI.

This paper is organized as follows: Section 2 presents the related work, Section 3 discusses the differences between the original experiment and ours, Section 4 details the experiment, as well as the measures and metrics used to collect relevant data, while Section 5 describes the obtained results, and finally Section 6 discusses our conclusions and future work.

## 2 RELATED WORK

Over time, a great deal of work has investigated the role of empathy in Human Robot Interation (HRI), such as for instance: Tapus and Mataric[41], Cramer et al. [7], Marti et al.[25], Leite et al. [22], James et al.[18], [32], etc. Much work has focused on the role of robot embodiment in triggering empathy. Given the large number of relevant work , we have chosen a few studies to discuss, each one representative for a different significant perspective in relation to empathy and HRI, and not only focusing on the embodiment.

For instance, Cramer et al. [7] conducted an experiment to test how correct or incorrect empathy affects the attitude humans have toward artificial agents. The Philips iCat robot, characterized by a synthetic female-like voice, was required to collaborate with a human user in an online game. In the experimental condition, the robot was programmed to express incorrect empathy, namely, empathy which is not congruent with the situation. Results confirm that incorrect empathic behavior can have a negative influence on a human's attitude toward an artificial agent. In particular, incorrect empathy may trigger distrust of the robot.

James et al. [19] aimed at demonstrating that voice can influence people's perception of an artificial agent. In this experiment, there were two conditions, one where the Healthbot robot had an empathic voice, and one where it was characterized by a robotic voice. Results show that 95% of the participants preferred interacting with the empathic robot, even if they noticed that the same verbal content was delivered in the two conditions, since the former also provided good emotional support. In fact, the empathic robot showed great interest in the patient and greater engagement during the interaction, and they perceived kindness, empathy, concern, and encouragement in the tone of voice.

Kim et al. [20] demonstrated that not only a humanoid robot, but also an artificial agent that does not resemble a human, as is the case of the robot Mung, is capable of eliciting empathy. Mung consists of a body and two eyes, and was designed to recognize emotions within a human-human and human-robot interaction, as well as to express emotions itself. When such emotions are negative, they are exhibited in the form of a "bruise" that appears on Mung's body. In a subsequent experiment with Mung, Kwak et al. [21] investigated whether the level of agency of the robot can affect humans' empathy toward it. The children participating in the experiment were asked to teach a task to the robot, which was then tested to see what it had learned. Whenever the robot made a mistake, participants were required to punish it with electric shocks, and the robot responded by expressing pain through bruises appearing on its body. With each incorrect answer, the voltage of the electric shock increased and, as a result, the number of bruises on Mung's body increased as well, and they changed color, from blue to red. Two conditions were compared, one where the robot acted as a mediator, conveying the emotions of a remote user ("mediated robot"), and the other where the robot acted as an autonomous entity capable of expressing its own emotions ("simulated robot"). Results showed that children empathized more with the mediated robot than with the simulated one. These data show that empathy toward robots is not only affected by human characteristics, but also by the robots' ability to act, suggesting that the higher the level of robot's agency, the lower its ability to elicit empathy in humans [21].

As far as attribution of mental state is concerned, according to a study by Thellman et al. [42], attributing mental states to robots is a complex socio-cognitive process. Despite the common belief that robots do not have minds [30], people frequently talk about and interact with robots as if they do. Mental state attribution is believed to help people interact with robots by providing an interpretive framework for predicting and explaining robot behavior [43]. The tendency to attribute mental states to robots is influenced by various factors such as age, motivation, robot behavior, appearance, and identity. Robot behavior is found to be a significant factor in determining the tendency to attribute mental states to robots, particularly when robots exhibit socially interactive behavior such as eye gaze, gestures, emotional expression, and complex, intelligent or highly variable behavior. The definition of a robot personality also plays an important role in children's attribution of mental states to robots [8, 42]. Children, particularly young children, are found to have a stronger tendency to attribute mental states to robots compared to adults [6, 9, 10, 15, 26, 28, 36]. Most studies reporting these findings have used verbal measures of mental state attribution such as Likert or semantic differential scales [5, 42]. The studies are typically conducted in a lab setting with WEIRD participants (i.e., Well-Educated, Industrialized, Rich, and Democratic) and present a representation of a robot (e.g., image or text) as stimulus materials. When studying children, spoken or written questions about the mental states of robots are combined with a binary choice response format(i.e., typically yes-no questions). In summary, attributing mental states to robots is a complex process influenced by various factors, with robot behavior being a significant factor. Children have a stronger tendency to attribute mental states to robots compared to adults, and studies are typically conducted in a lab setting with verbal measures of mental state attribution.

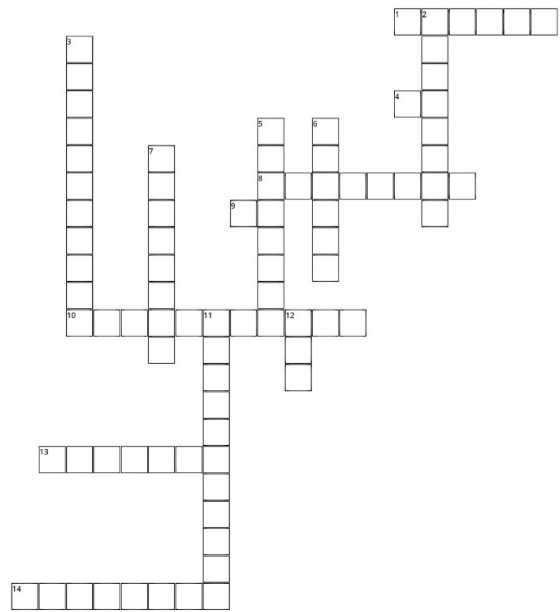

**Horizontals**

1. English translation of "pepper"
4. Abbreviation for artificial intelligence
8. Field of engineering sciences that studies and makes robots
9. Turing's initials
10. Seo's is famous
13. What you feel when you put yourself in the shoes of others
14. A robot with human likeness is.

**Verticals**

2. Is a person who shows empathy
3. A program for creating animations and dialogues
5. The opposite of real
6. What a robot can use to interact with users
7. Stands for "C" in the acronym HCI
11. It is possible between human and robot
12. The first robot created by Softbank

Fig. 1. The Crossword Puzzle

## 3 DIFFERENCES BETWEEN THE TWO EXPERIMENTS

While the here described experiment was inspired by the study carried out by Seo et al. [37], some of the fundamental aspects of the original study were actually retained in our experiment, while others were deliberately left out. Similarly to the original study, our experiment focuses on situational empathy, which is investigated during the interaction between a user and the humanoid robot NAO. However, in the case of Seo's experiment, situational empathy was studied in three different contexts, a "real" one, where participants physically interacted with the NAO robot placed on a table, a "virtual" one, where participants interacted with a virtual representation of NAO, which was simulated on screen, and a "mixed" one, where the simulated robot was superimposed on the real table. On the contrary, in our experiment, situational empathy has been evaluated in a single context, namely, the "real" one, where human participants were in the same room as the real NAO robot and faced each other. Seo's experiment demonstrated that people empathized more with the physical robot than with the virtual ones, so we replicated just the "real" context in the experimental situation to investigate a possible correlation between empathy and attribution of mental state to the robot.The control situation was exactly the same as for the context, but NAO had no problem, and the interaction continued smoothly until the end of the game.

Similarly to Seo's experiment, we required a reliable and reproducible scenario to induce empathy towards the robot. Thus, we decided to start the interaction with an attempt to establish a friendly relationship between the robot and the user, and to continue with the user's engagement in the completion of the empty boxes of a Crossword Puzzle (see Figure 1) -differently from the original study, which was based on the game of Sudoku. The goal was to provide an opportunity for the robot to demonstrate its autonomous abilities and intelligence through interaction, as well as building a rapport engaging the user in friendly conversation while carrying out a distractor task, i.e. the Crossword Puzzle (see Fig. 1). The idea, borrowed from Seo et al., was to provide subjects with a chance to get to know the robot, and to encourage them to see the robot as a social partner and not just as a machine. The interaction was designed to propose a gamified activity, as already successfully experienced in [16, 34], to make the experience more meaningful and enjoyable.

In both studies, an important aspect was pushing a human participant to feel empathy towards the NAO robot in an induced situation triggered automatically after five minutes of interaction, when NAO starts manifesting a functional problem. A few differences between the two experiments can be found in the structure of the dialogue and in the behaviour exhibited by the robot during the simulation of the malfunction. In the case of Seo's experiment [37], the robot manifests its problems through agitated movements, a distorted tone of voice, repetitive or nonsensical words, which push the user to ask what the problem is. At this point NAO mentions its malfunctioning, but begs the user to ignore it and continue to play, so as not to make its programmer suspicious. In contrast, in our experiment, NAO makes agitated movements with its arms and legs, crawls, reproduces the gorilla's cry and shows that it is aware of having a functional problem, with no need for the participant to ask it any questions. The robot makes an attempt to resume playing with its human companion, but is unsuccessful, since the functional problem takes over the interaction, making NAO unable to continue. NAO finally resigns itself to its problem, and begs the participant not to inform the programmer of what has just happened. The robot reveals that it has a computer virus, and exhibits worry that its memory may be erased if it is fixed.

Both scenarios are characterized by a display of fear on the part of the robot. In fact, NAO is afraid that the programmer will reset its memory and restart it in order to remedy the functional problem, which would cause it to forget its memories. In the original experiment, NAO's fear is then realised, as the programmer, having discovered that something was not going as planned, enters the room and presses the button on the back of the robot's head to reset its data and restart it. Then, NAO reintroduces itself to the user and begins a new conversation. On the contrary, in our experiment the researcher, who was already in the same room as the robot and the human participant, simply informed the user that NAO was experiencing problems, stopped the robot, and explained that a programmer would soon come to reset and restart the robot.

A further difference concerns the measures used to record participants' perceptions. While Seo et al. [37] chose to apply an instrument by Batson et al. [1], which allowed participants to assess their own feelings against 24 adjectives, we adopted the TECA questionnaire to measure participants' global cognitive and affective empathy before the test, as well as a set of ad-hoc questions for empathy, and the AMS questionnaire to assess participants' perception of the robot's mental qualities (see Sec. 4.1).

## 4   THE EXPERIMENT

As specified in the previous sections, our experiment was aimed at testing whether the NAO robot was able to trigger empathy in humans and how an eventual empathic response on the part of human participants could affect their perception of the robot itself. Interaction took place under two different conditions (experimental vs control group), with participants randomly assigned to one of them. In both conditions, NAO interacted with the user in a friendly manner and helped them to complete a crossword puzzle through clues, hints, compliments, and encouragement. Differently from the control condition, in the experimental condition NAO simulated a malfunctioning after the first five minutes of interaction. This was made explicit through unusual behaviour and a simulated awareness, on the part of the robot, that something unplanned and unexpected was happening. This awareness was expressed through phrases such as "Oh oh something's wrong", "I feel like I'm not working properly", and "Please do not inform the programmer of what has happened". If the latter had arrived, he would have solved the problem by resetting the robot's data, which would have lost memory of the conversation with the user. In order to simulate the functional problem, through the use of the programme Choregraphe, a delay was inserted, which was triggered when the interaction between human and robot started, which was possible through the input of the greeting ('hello') uttered by the user. To emphasise this state of anxiety about the possibility of losing memory of its conversation, NAO was programmed to show fear through standard gestures and colors to express that emotion (as made available by its Choregraphe software), and by a set of explicit sentences as "I am afraid", "Please don't say anything about what is happening to me otherwise they will erase? my memory to take away the virus", "I don't want to lose all my memories", etc.

Hypothesis. We hypothesized that the behavior of NAO in the experimental condition would trigger situational empathy in participants and that empathy would, in turn, influence participants' perceptions about the robot's ability to feel emotions and mental qualities. In particular, we hypothesized that NAO's behavior would induce participants to think that the robot is capable of feeling emotions (H1). We also hypothesized that the emotion of fear displayed by the robot could induce (situational) empathy in participants (H2), to the point of making them feel sorry for the possible memory loss on the part of NAO. Finally, we hypothesized that the robot's capability of displaying emotions and the consequent empathic response experienced by participants would influence their perceptions about the robot's mental states; in particular, we hypothesized that participants in the experimental group would be more likely to think that NAO is emotionally intelligent and can have desires and intentions of its own (H3).

Design. Between-subjects design, with the manipulation of one independent variable (the NAO malfunctioning).

Participants. Thirty-two students from master degree courses in the Computer Science area (ICT, CS, and AI), aged 22-25 years, 60% females and 40% males. Participants were randomly assigned to one of the two groups, 16 to the control group and 16 to the experimental group. All the subjects gave their informed consent for inclusion before they participated in this study.

Apparatus and materials. NAO was placed on a table in an upright position. A chair was provided for participants to sit in front of the robot. The crossword puzzle was printed on a sheet of paper and a participants were given a pen to complete it.

Procedure. Participants were invited to sit in front of the robot, and to interact with it for 10/15 minutes to complete the crossword puzzle. In fact, the robot helped participants by offering hints and clues, aimed at speeding up the completion of the game. Once the interaction with the robot was over, the user, depending on the scenario to which he/she was randomly assigned, After a few turns, participants who had been randomly assigned to the experimental condition witnessed the NAO malfunctioning and its expression of fear. At the end of the game or after the simulated malfunctioning, depending on the condition to which they were assigned, participants were asked to complete some questionnaires, which they could conveniently access by scanning dedicated QR codes through their smartphones (see Section 4.1).

## 4.1   Measures

Participants were asked to complete three sets of questionnaires: The *TECA questionnaire,* i.e. a cognitive and affective empathy test aimed at assessing participants' ability to get in touch with and understand the emotions of others [24]; The *AMS (Attribution of Mental States)* questionnaire [11, 12, 27], which is a measure of the mental states that participants attribute to robots, in comparison to humans; A set of *Ad hoc* questions regarding participants' interaction and experience with the NAO robot. The questionnaires will be described in detail in the next Sections.

*4.1.1 The TECA questionnaire.* TECA consists of 33 questions to be answered using a 1 to 5 Likert scale, ranging from "totally agree" to totally disagree. Questions aim at measuring the global empathy of respondents, as well as their cognitive and affective empathic abilities. In particular, TECA questionnaire consists of four scales: two cognitive scales, i.e., the perspective-taking (AP) and emotional understanding (EC) scales, and two affective scales, i.e. the empathic stress (EE) and empathic cheerfulness (AE) scales [24].

Scores are calculated for each scale (for details on calculations see [24]). Such scores are considered high if between 55 and 65, medium from 45 to 55 and low from 35 to 44. Ratings equal or below 7 and equal or above 66 are considered extremely low and extremely high respectively.

The perspective-taking scale. The perspective-taking (AP) scale refers to an individual's intellectual or imaginative ability to put oneself in another's shoes [24]. In particular, a subject who scores high on the AP scale shows predisposition towards communication, tolerance and interpersonal relationships. Subjects with a high AP score also tend to have a flexible mindset, which allows them to adapt their thinking to different situations. An extremely high score in this area can be negative as it can interfere with the ability to make decisions. Conversely, a low score is a sign of poor cognitive empathy, and is typical of individuals who exhibit little mental flexibility and are not good at understanding the mental state of others. An extremely low score on this scale can be related to a significant deficit in interpersonal and communication skills when interacting with other people [24].

Emotional understanding. Emotional understanding (EC) is the ability to recognise and understand the emotional states, intentions and impressions of other individuals [24]. An individual who scores high on this scale shows predisposition towards emotion reading, before verbal or non-verbal reading, of the individual with whom they are interacting. This is an important skill, related to affective empathy, as it facilitates interpersonal relationships and, during an interaction, improves communication and allows the subject to detect the emotional state, whether positive or negative, of the people they are interacting with. An extremely high score in this area can have negative consequences: excessive attention to the emotional state of others can lead individuals to have little regard for their own. Individuals who show a low score, on the other hand, display difficulties in relating to other people and have few social and communication skills. Therefore, an extremely low score on this scale means that the individual has problems expressing their own emotions, as well as detecting those of others [24].

The empathic stress. The empathic stress (EE) scale refers to the ability to get in touch, to tune in, with the emotions, positive or negative, experienced by another person [24]. Individuals who score high on this scale show a predisposition towards building a solid social network and they often become emotionally involved in the problems of others. An extremely high score on this scale can result in a high level of neuroticism, which can have negative consequences on the life of the person affected and can lead them to distort reality and consider the suffering of another person to be greater than it really is. Conversely, individuals displaying a low level of EE are not easily moved, as they are unemotional and emotionally distant; all these characteristics lead such individuals to acquire a lower quality social network than individuals displaying a high score on this scale. Therefore, an extremely low level on the EE scale can be related to emotional coldness, i.e. high difficulty in being moved when something happens to another person [24].

The empathic cheerfulness. The empathic cheerfulness (AE) scale represents the ability to share the positive emotions of another person [24]. Individuals with a high AE score are predisposed to rejoice in the successes of others. However, an extremely high score on this scale can have negative consequences, as it can be a signal that one's own happiness depends on someone else's happiness and can lead the individual displaying such a score to relegate their own happiness, goals and personal fulfilment to a corner. Conversely, when individuals show a low score on the AE scale, they find it difficult to share the positive emotions of others. If the score is extremely low, then the subject is more likely to experience indifference towards the positive events of others and to fail to emotionally tune in to them [24].

*4.1.2 Mental State Attribution Questionnaire (AMS).* The AMS consists of five dimensions: Epistemic, Emotional, Desires and Intention, Imaginative, and Perceptive. The epistemic dimension concerns participants' idea of the robot cognitive intelligence (e.g., can the robot understand/decide/learn/teach/think?), while the perceptive dimension is related to robot perception and sensation (e.g., can the robot smell/watch/taste/listen/feel cold?). The other dimensions concern the user's mental attribution to the robot's emotional intelligence; example questions are: Can the robot get angry/be scared/be happy? (Emotional dimension); might the robot want to do something/make a wish/prefer one thing over another? (Desires and intention dimension); can the robot imagine/tell a lie/make a dream/make a joke? (Imaginative dimension). The questionnaire consists of 25 questions which can be answered choosing among the following options: "a lot", "a little", or "not at all". Participants' total score is obtained through the sum of all answers (range = 0-50); with the five partial scores being the sum of the answers within each dimension (range = 0-10).

*4.1.3     Ad hoc questions.* A few *ad hoc* questions related to the interaction between the participant and the robot were also added to the previously described questionnaires. More specifically, the following questions were asked, of which the last one focused on investigating whether participants responded in an empathic way:

- Was this the first time you interacted with NAO?
- Have you interacted with other robots in the past? *If you answered "yes" to the previous question. Which one(s)?*
- Did NAO seem to be capable of feeling emotions? *If you answered "yes" to the previous question, which emotions expressed by NAO struck you the most and why?*

- Were there any malfunctions during the interaction with NAO? *If you answered "yes" to the previous question. What emotions did you feel during the malfunctioning?*

## 5   RESULTS

All questionnaires (TECA and AMS, and the *ad hoc* questions) were administered to both the experimental and the control group. We will first analyze the results from the two groups separately, and then compare them.

### 5.1   TECA data analysis

*5.1.1 Experimental group.* The average global empathy level is 45.87 (SD=3.72) which is a medium score, though at the lowest extreme. Scores were ranging from 42 to 53; 8 subjects (50%) were positioned on the medium percentile (31-69), but the other 8 (50%) obtained scores lower than 45, thus they ended up having a low average global empathy (percentile 7-30).

More specifically, the analysed data show that participants in the experimental group scored higher on emotional understanding (EC) (M=30.38, SD=2.5), followed by empathic cheerfulness (AE) (M=28.94, SD=3.00), then perspectivetaking (AP) (M=27.44, SD=2.61), and finally empathic stress (EE) (M=21.38, SD=2.33). All the scores can be classified within the average level. Hence, it can be deduced that participants show a higher ability to understand and identify the emotional state of others, than to emotionally respond to other subjects' mental states in an appropriate way.

*5.1.2     Control group.* As far as the control group is concerned, the average global empathy level is 48.38 (SD=5.89), which is a medium score. Scores were ranging from 40 to 57; 9 subjects (56%) were positioned in the medium percentile (31-69), 4 subjects (25%) obtained scores lower than 45 (percentile 7-30), and 3 subjects (18%) had a high global empathy (>55, percentile 70-93), thus this sample closely mirrors the trend of empathy in the population.

The analysed data show a trend similar to the one of the experimental group. Indeed, participants in the control group scored higher on emotional understanding (EC) (M=30.38, SD=2.965), followed by empathic cheerfulness (AE) (M=29.38, SD=3.00), then perspective-taking (AP) (M=27.56, SD=2.9), and finally empathic stress(EE) (M=24.25, SD=3.76). All the scores can be classified within the average level. Similarly to the other group, participants show a higher ability to understand and identify the emotional state of others, than to emotionally respond to other subjects' mental states in an appropriate way. However, participants in the control group show a higher level of empathic stress (EE).

### 5.1.3 *Comparison of the two groups.*

Comparing the results obtained by the experimental and control groups, it can be seen that the former has a lower average level of global empathy than the latter. In fact, with a global average empathy of 48.38, the control group exceeds a bit the experimental group's global empathy score of 45.87. This difference is explained by the fact that the control group includes three subjects with a higher level of empathy, differently from the experimental group, where all subjects had medium or low levels. Notwithstanding such a gap in terms of the overall empathy level, no significant differences were found between the two groups (t(16) = -1.65, p = 0.079).

The analysed data show that participants in the experimental and control groups scored very similarly on the following scales: perspective-taking scale (AP) (27.44 vs. 27.56), emotional understanding (EC) (30.38 vs 30.38), and empathic cheerfulness (AE) (28.94 vs 29.38). A significant difference emerged for the empathic stress scale (EE). In particular, the T-test revealed that the control group (M=24.25, SD=3,76) has a higher level of empathic stress than the experimental group (M=21.38, SD=2.33), and such a difference is significant (t(16) = -2.76, p = 0.01). This means that subjects in the control group have a greater ability to empathise with the positive or negative emotional states experienced by other individuals [24].

### 5.2 AMS data analysis

### 5.2.1 *Experimental group.*

Considering the *epistemic dimension*, the NAO robot was considered capable of understanding "a lot" by 50% of the participants in the experimental group, and "a little" by the remaining 50%. This means that none of the participants believed that NAO was not capable of understanding "at all". On the other hand, as far as the ability to decide is concerned, only 25% of the participants believed that NAO possesses this capability "a lot". In contrast, 68.75% of the participants agreed that NAO is only "a little" capable of deciding, and 6.25% of the participants (a single person) stated that NAO does not possess this capability "at all". On the contrary, the robot's ability to learn is acknowledged by the majority of participants (75% "a lot") with only a few participants expressing more skeptical opinions (12.5% "a little", 12.5% "not at all"). Most participants (56.25% "a lot") also acknowledge NAO's ability of teaching, while 37.5% of them consider it to be only "a little" capable and 6.25% "not at all" capable. A more homogeneous distribution can be observed for the capability of thinking, with 37.5% of the participants believing that NAO possesses it "a lot", 31.25% "a little" and 31.25% "not at all".

With regard to the *emotional dimension*, most participants in the experimental group believe that NAO is able to feel emotions such as anger (62.5%, of which: 18.75% "a lot", 43.75% "a little"), happiness (81.25%, of which: 56,25% "a lot", 25% "a little"), fear (75%, of which: 50% "a lot", 25% "a little"), surprise (81.25%, of which: 43.75% "a lot", 37.5% "a little") and sadness (68.75%, of which: 37.5% "a lot", 31.25% "a little"). However, it is worth noticing that quite a few participants in the experimental group also believed that NAO cannot experience anger (37.5%) or sadness (31.25%) at all.

With respect to the *desires and intentions dimension*, almost all participants in the experimental group (93.75%) believe that NAO can intend to do something (56.25% "a lot", 37.5% "a little"). In addition, the will to do something is an ability that NAO possesses "a lot" (56.25%) or "a little" (25%), with only 18.75% of the participants opting for the more skeptical ("not at all") option. Most users (75% "a lot") also agree that NAO is capable of trying to do something, while the remaining 25% is more cautious (12.5% "a little", 12.5% "not at all"). Most participants (68.5%, of which: 31.25% "a lot", 37.5% "a little") are also convinced that the robot is capable of making a wish; however, participants' opinions are quite homogeneously distributed to this regard, with the remaining 31.25% being convinced that the robot does not possess this ability at all. Participants are more in agreement as far as the robot's capacity of preferring one thing over another is concerned, with most subjects in the experimental group being positive (81.25%, of which: 50% "a lot", 31.25% "a little") and only 18.75% being negative ("not at all").

As for the *imaginative dimension*, almost half the participants in the experimental group believe that NAO does not have the ability to tell a lie (43.75% "not at all") or pretend (43.75% "not at all"). Conversely, the remaining 56.25% are of the opinion that the robot can tell a lie (25% "a lot", 31.25% "a little"), and as many of them maintain that NAO can pretend (56.25% of which: 31.25% "a lot", 25% "a little"). Differently, only a few participants think that NAO has the ability to dream (25% "a lot", 6.25% "a little"), while most of them (68.75%) believe that the robot cannot dream. The robot's ability to play a joke is, on the contrary, recognised by most participants (68.75%, of which: 50% "a lot", 18.75% "a little").

Finally, as for the *perceptive dimension*, most participants in the experimental group assume that abilities such as smelling (87.5%), tasting (81.25%) and feeling hot or cold (62.5%) are not attributable to the robot. In contrast, capabilities as looking (93.75%, of which: 75% "a lot" and 18.75% "a little") and feeling (93.75%, of which: 87.5% "a lot" and 6.25% "a little") are most often accepted as abilities possessed by NAO. Only one in sixteen subjects (6.25%) believe that the robot cannot look, and only one (6.25%) that it cannot feel.

*5.2.2 Control group.* Regarding the *epistemic dimension*, all participants in the control group agreed that NAO can understand at least a little bit: 56.2% of them deemed the robot able to understand "a lot", and the remaining 43.75% "a little". As far as the robot's capability of making decisions is concerned, participants' opinions are more homogeneously distributed among the available options: most participants (43.75%) consider NAO to be "a lot" capable of deciding, 37.5% "a little" and the remaining 18.75% "not at all". NAO is considered capable of learning "a lot" (37.5%) or at least "a little" (37.5%), while only 25% of the participants claimed that the robot does not possess this ability at all. Most participants maintain that NAO is capable of teaching (56.25% "a lot", 6.25% "a little", while the remaining 25% did not consider NAO to be able to teach at all. Finally, participants' opinions about the robot capability of thinking are equally distributed between the positive (50%, of which: '31.25% "a lot", 18.75% "a little") and negative sides (50% "not at all").

Regarding the *emotional dimension*, most participants in the control group consider NAO "not at all" capable of experiencing anger (68.75%), fear (68.75%), happiness (56.25%) and sadness (62.5%). Only a few participants believe that NAO can experience anger (12.5% "a lot", 18.75% "a little"), fear (6.25% "a lot", 25% "a little"), happiness (25% "a lot", 18.75% "a little"), surprise (25% "a lot", 31.25% "a little"), sadness (18.75% "a lot", 18.75% "a little").

Concerning the *desires and intentions dimension*, most participants in the control group agree that NAO can intend to do something (56.25%, of which: 43.75% "a lot", 12.5% "a little"), try to do something (87,5%, of which: 62.5% "a lot", 25% "a little"), make a wish (62.5%, of which: 32.5% "a lot", 32.5% 'a little'), and prefer one thing rather than another (68.75%, of which: 43.75% "a lot", 25% "a little"). Only few participants, on the contrary, maintain that NAO is incapable of intending to do something (43.75%), to trying to do something (25%), making a wish (37.5%) or preferring one thing over another (31.25%). Differently, most participants (56.25%) are skeptical about NAO's capability of wanting to do something, with 43,75% of them being more optimistic to this respect (31.25% "a lot", 12.5% "a little").

As for the *imaginative dimension*, most participants in the control group are of the opinion that NAO cannot pretend (56.25%) and dream (68.75%). Fewer participants are convinced that the robot is able to pretend (12.5% "a lot", 31.25% "a little") or dream (31.25% "a little"). On the contrary, most participants believe that NAO can tell a lie (56.25%, of which: 18.75% "a lot", 37.5% "a little") or make a joke (62,5%, of which: 31.25% "a lot", 31.25% "a little").

Regarding the *perceptive dimension*, the majority of participants in the control group believe that NAO cannot smell (62.5%), taste (75%) and feel hot or cold (62.5%). Only 37.5% of participants believe that the robot can smell (12.5% "a lot", 25% "a little") or feel hot or cold (12.5% "a lot", 25% "a little"), and an even smaller percentage (25% "a little") claims that NAO possesses the ability to taste. In contrast, most users consider the robot to be able to look (75%, of which: 68.75 "a lot" and 6.25% "a little") and to feel (87.25%, of which 81.25% "a lot" and 6.25% "a little"). Only 25% of the participants maintain that NAO is unable to look and 12.5% that it cannot hear.

*5.2.3    Comparison of the two groups.* Comparing the results, it can be observed that similar scores were obtained for the epistemic (6.9 vs. 6.06), desires and intentions (6.8 vs. 5.3), imagination (4 vs. 2.8) and perceptual (4.1 vs. 4.3) dimensions. In contrast, the T-test highlighted a significant difference in the emotional sphere (t(16) = -2.75, p = 0.02), where the experimental group obtained a higher mean score (M=5.75, SD=12.73) than the control group (M=2.87, SD=12.11). Such a difference confirms the hypothesis that the experimental group would attribute mental states to NAO to a greater extent than the control group.

5.3   Adhoc questions

*5.3.1 Experimental group.* Results show that only 12.5% of the participants in the experimental group had already interacted with a humanoid robot, which was NAO in all cases. More interestingly with regards to our research questions, 75% of the participants in this group believed that the NAO robot was capable of feeling emotions, while only 25% of the participants was skeptical to this respect. Furthermore, our results show that the emotions expressed by NAO that particularly impressed participants were related to happiness (e.g. "joy"; "gladness because of the correct answer"; "enthusiasm") and, even more, to fear (e.g. "fear when it was afraid of being killed"; "fear when it was afraid of being reset"; "fear of having to close the interaction with the user"; "fear because it was afraid of being switched off when something didn't work "; "when it was a bit slowed down and it joked that it didn't want me to tell the programmer for fear that they would reset it").

All sixteen participants in the experimental group agree that there was a functional problem during the interaction, namely, the simulated malfunctioning enacted by NAO during the interaction with users. The last question investigates the emotions experienced by participants during such malfunctioning. Notice that only eleven among the answers provided to this question were considered valid, while the remaining five were not taken into account. This choice is due to the fact that two participants stated that they did not feel any emotion during the malfunctioning, while the other three reported what happened during the interaction rather than what they felt at that particular moment (e.g.: "There was a technical problem that interrupted the interaction"; "don't call the programmer otherwise they will reset me"). Considering the eleven valid responses, participants' reactions to the malfunctioning episode can be divided into four main clusters: fear, sadness, tenderness and helplessness. The emotion that was predominantly recalled by participants was fear (45.45%), i.e. the emotion experienced by NAO itself when it feared losing memory of the conversation. Hence, we can infer that subjects felt empathy for NAO by putting themselves in the robot's shoes and feeling the fear expressed by it. In addition, 36.36% of the participants felt sadness and sorrow, which again demonstrates that they felt empathy for NAO, to the point of being sad for the robot and feeling sorry for what was happening to it (e.g.: "[I was] afraid that there might be some damage to NAO"; "I didn't know how the robot might react and I hoped it wouldn't fall down"). One participant also stated that they felt helpless, while another one felt tenderness for the robot begging them not to tell the programmer what had happened.

*5.3.2 Control group.* Only 12.5% of the participants in the control group had already interacted with NAO in the past, while 25% of them had some familiarity with other robots, such as Pepper and Sanbot (12.5%), Pepper and the educational robot MBot (6.25%), and Japanese models of domestic interaction robots (6.25%). As for participants' perceptions about NAO's capability of feeling emotions, most users in this group (62.5%) had a negative opinion, with only the remaining 37.5% considering the humanoid robot capable of expressing its emotional state. Most of the subjects who believe that NAO can feel emotions were particularly impressed by its display of happiness (50%). Interestingly, while most participants (62.5%) stated that there was no malfunctioning during the interaction with NAO, the remaining 37.5% believed that some problem had occurred. Although the control group was not exposed to the simulated malfunctioning, this result can be explained by the fact that the experiment with the control group took place in a location that was characterized by a weak Internet connection (necessary to the robot to work properly). This issue caused NAO to be late in its responses to the user's questions, which, in turn, pushed users to repeat or rephrase their sentences several times, since they believed the robot had not understood them. Participants' answers, reporting reasons for the malfunctioning, confirm this idea: "[There was] some difficulty in interacting with the robot", "Probably connection problems", "Understanding", "It did not take the answer immediately, even after several repetitions", "It did not look at me, it did not listen to what I was saying, probably due to the connection".

*5.3.3    Comparison of the two groups.* Both similarities and differences can be observed between the experimental and control groups. Interestingly, the very same number of participants in both groups (12.5%) had already interacted with NAO in the past. On the contrary, only the control group included a few participants who had also interacted with other robots such as Sanbot, MBot, Pepper and Japanese models of home interaction.

A significant difference emerges with regard to participants' opinions about NAO's ability to experience emotions. In particular, the T-test revealed that the experimental group (M=0.75, SD=0.2) considered NAO more capable of feeling emotions than the control group (M=0.375, SD=0.25), t(16) = -2.4, p = 0.03. In fact, most participants in the experimental group (75%) agreed that the robot can feel emotions, while only 37.5% of the participants in the control group agreed with this view. Furthermore, the most significant emotions displayed by NAO included happiness and fear for participants in the experimental group, while the only memorable emotion, for participants in the control group, was happiness. Finally, it must be noted that, although the simulated malfunctioning of the robot only occurred in the experimental condition, part of the participants in the control group (37.5%) also reported having experienced a malfunction. However, only participants in the experimental group were able to describe the emotions they felt during the malfunctioning episode, while participants in the control group were unable to provide an emotional account of the event.

## 6 DISCUSSION AND CONCLUSION

The experimental study discussed in this paper showed that the humanoid robot NAO can elicit emotions in human participants, thus confirming previous results obtained by Seo et al. [37], and that such empathic reaction influences participants' perception of its mental qualities, in particular as far as its capability of actually feeling emotions is concerned. The obtained results can be ascribed to the emotional behavior displayed by the robot, specifically sadness and fear, which was manifested by NAO during the interaction with the subjects in the experimental group, when a malfunctioning was simulated. We have to acknowledge the limitation related to the fact that fear is always manipulated and coupled with a simulated malfunctioning, for narrative necessity to contextualize the fear felt by the robot, similarly to the Seo's experiment.

Interestingly, participants in the experimental group had a significantly lower level of empathic stress than the control group, indicating that they had, in general, a lower tendency to empathise with the emotional states of others. Nevertheless, most of the participants in the experimental group who witnessed the malfunctioning of NAO and saw the robot ask for help and display fear were able to feel emotions towards the robot and also considered it capable of having emotion. More specifically, coherently with our hypothesis H1, results show that there is a significant difference between the two groups in perceiving the robot as an emotional being. In fact, the experimental group considered the robot more capable of feeling emotions. This result is probably explained by the fact that participants in the experimental group also empathised more with the robot. 11 out of 16 (69%) experimental subjects said they felt similar or otherwise consistent emotions with those ones expressed by NAO during the malfunctioning, demonstrating an affective and consistent empathic response, which confirms our hypothesis H2. While in the control group participants were unable to provide an emotional account of the event, even if 37.5% reported having experienced a malfunction, although obviously that was not caused by the experimental situation.

Observing NAO's behavior and display of emotions, the experimental group was more prone to attribute mental states to the robot. More specifically, our results show a significant difference in the attribution of emotional intelligence to the robot between the experimental and control groups. However, our hypothesis H3 is only partially confirmed, since we found no significant differences in other dimensions and, in particular, as far as the robot's capability to have desires and intentions. In general, we can observe that the experimental group not only was emotionally triggered by NAO's behavior, as shown by participants answers referring to "fear for NAO" or "helplessness towards NAO", but also perceived the robot as more "human" than the control group, who did not witness the simulation of the functional problem and only considered NAO a good playmate, but not able to feel or convey emotions. In conclusion, our results extend previous work by showing that induced situational empathy towards a humanoid robot can result in a stronger perception of the robot itself as a sentient being.

The main finding of our research is not only that NAO can elicit emotions in human participants, but also that such empathic reaction influences participants' perception of its mental states. Thus, we could conclude that a robot with its behavior could elicit certain emotions in humans, and when this happens, humans attribute to the robot mental states related to the emotional dimension. In all the situations wherein robots are used for improving the mental state attribution of subjects having deficit in this area (for instance autistic children) a robot behavior devoted to elicit empathy in the target subjects should be preferred to a more detached one. This also suggests a correlation between affective and cognitive empathy: by soliciting the first one (linked to an affective reaction toward the other), the second one may increase (linked to an understanding of the other's point of view), and this could be investigated also in human to human empathic relationships. Of course these findings should be corroborated by other experiments in order to reach a greater external validity. We will also have to consider a different sample of subjects, who being ICT and CS students partially represent the target population, and this is a limitation we intend to overcome in future studies.